\documentclass[11pt]{article}

\usepackage[preprint]{acl}

\usepackage{times}
\usepackage{latexsym}
\usepackage[T1]{fontenc}
\usepackage[utf8]{inputenc}
\usepackage{microtype}
\usepackage{inconsolata}
\usepackage{graphicx}
\usepackage{amsmath}
\usepackage{amssymb}
\usepackage{booktabs}
\usepackage{multirow}
\usepackage{float}
\usepackage{hyperref}

\title{Polarization Detection: A Hybrid Approach with AfroXLMR-Social and DeBERTa for Low- and High-Resource Settings}

\author{Muhammad Abdullahi Said \\
  African Institute for Mathematical Sciences (AIMS) \\
  \texttt{mohdasaid@aims.ac.za} \\}

\begin{document}
\maketitle

\begin{abstract}
The rapid proliferation of online polarization threatens social cohesion, necessitating robust automated detection systems that operate effectively across diverse linguistic contexts. This paper presents our system description for the POLAR Shared Task 2026, focusing on the detection and characterization of polarized discourse in English and Hausa. We propose a hybrid modeling strategy: for English binary detection, we leverage the monolingual strength of \textbf{DeBERTa}, while for Hausa and all fine-grained subtasks (Types and Manifestations), we utilize \textbf{AfroXLMR-Social}. This domain-adapted multilingual model proved critical for capturing the nuances of polarization in social media text. To further address computational constraints and data scarcity, we implement Low-Rank Adaptation (LoRA) and textual data augmentation via \texttt{nlpaug}. We report competitive results across all three subtasks, demonstrating that model selection tailored to specific subtask requirements yields the best balance of performance.
\url{https://github.com/mohdasaid/NLP-LLM/blob/main/Polarization.ipynb}
\end{abstract}

\section{Introduction}

Online polarization defined as the sharp division of public opinion into opposing factions characterized by hostility and lack of empathy has become a pervasive issue in digital discourse. The POLAR Shared Task 2026 aims to benchmark Natural Language Processing (NLP) systems on their ability to detect this phenomenon across a wide range of languages and cultures \citep{naseem2025polar}.

Detecting polarization extends beyond traditional sentiment analysis. It requires capturing subtle rhetorical strategies such as "us vs. them" framing, invalidation, and dehumanization. This complexity is exacerbated in low-resource languages like Hausa, where labeled data is scarce and standard multilingual models often lack sufficient exposure to dialectal variations.

In this work, we propose a strategic, hybrid training pipeline. Our key contributions are:
\begin{enumerate}
\item Task-Specific Model Selection: We observe that while English benefits from specialized monolingual models for binary detection (Subtask 1), the fine-grained tasks (Subtasks 2 \& 3) in both languages require the robust social-media understanding provided by AfroXLMR-Social\citep{belay2025afroxlmr}.
\item Domain-Adaptive Efficacy: We demonstrate that AfroXLMR-Social, a model continued-pre-trained on African social media content, is indispensable for the Hausa subtasks, significantly outperforming generic multilingual baselines.
\item Parameter Efficiency \& Augmentation: We employ LoRA (Low-Rank Adaptation) \citep{hu2021lora} to fine-tune effectively on consumer hardware and use \texttt{nlpaug} \citep{ma2019nlpaug} to generate synthetic examples, addressing severe class imbalance in the manifestation labels.
\end{enumerate}

\section{Related Work}

\subsection{Polarization Detection}
Early computational approaches to polarization relied on network topology, inferring ideological stance from retweet or follower graphs. However, content-based approaches using Transformers have recently gained traction. Studies have shown that models like BERT and RoBERTa can detect stance and toxicity, which are proxies for polarization. The POLAR benchmark advances this by explicitly taxonomizing polarization into types (political, religious) and rhetorical manifestations.

\subsection{Multilingual Models for African NLP}
Massively multilingual models like mBERT and XLM-R often underperform on African languages due to the "curse of multilinguality," where capacity is diluted across too many languages. \textbf{AfroXLMR-Social} \citep{belay2025afroxlmr} builds upon the success of the original AfroXLMR \citep{alabi2022adapting} by incorporating "Domain-Adaptive Pre-training" (DAPT) on social media corpora. This adaptation is crucial for Hausa, as it aligns the model's internal representations with the noisy, informal, and code-mixed text styles found in the POLAR dataset.

\subsection{Efficient Fine-Tuning}
As model sizes grow, full fine-tuning becomes prohibitively expensive. Parameter-Efficient Fine-Tuning (PEFT) methods like Adapters and LoRA have emerged as standard solutions. LoRA, in particular, has been shown to match or exceed full fine-tuning performance in low-data regimes by reducing the risk of catastrophic forgetting. By freezing the pre-trained backbone, we maintain the linguistic knowledge acquired during the extensive DAPT phase.

\section{Task Description}

We address three hierarchical subtasks provided by the organizers:
\begin{enumerate}
    \item \textbf{Subtask 1 (Detection):} A binary classification task to determine if a text $x$ contains polarized content ($y \in \{0, 1\}$.
    \item \textbf{Subtask 2 (Type):} A multi-label classification task identifying the topic of polarization. Labels include \textit{Political}, \textit{Religious}, \textit{Racial/Ethnic}, \textit{Gender/Sexual Identity}, and \textit{Other}.
    \item \textbf{Subtask 3 (Manifestation):} A multi-label task identifying rhetorical devices used to express polarization. Labels include \textit{Vilification}, \textit{Stereotyping}, \textit{Dehumanization}, \textit{Extreme Language}, \textit{Lack of Empathy}, and \textit{Invalidation}.
\end{enumerate}

\section{Methodology}

\subsection{Model Architecture Strategy}

\subsubsection{English Subtask 1: DeBERTa}
For the binary detection task in English, we utilized \textbf{DeBERTa-v3-base}. DeBERTa improves upon BERT and RoBERTa by using disentangled attention and an enhanced mask decoder, making it highly effective for subtle classification tasks in high-resource languages where standard grammar prevails.

\subsubsection{Hausa \& Subtasks 2/3: AfroXLMR-Social}
For all Hausa tasks and the fine-grained English tasks (Types and Manifestations), we employed \textbf{AfroXLMR-Social}. The complexity of Subtasks 2 and 3 requires a model that understands the social context of language slang, hashtags, and informal phrasing which is the core strength of this domain-adapted checkpoint.

\subsection{Low-Rank Adaptation (LoRA)}
To adapt these models efficiently, we freeze the pre-trained backbone and inject trainable low-rank matrices ($A, B$) into the Query ($Q$) and Value ($V$) attention projections.
\begin{equation}
    h = W_0 x + \frac{\alpha}{r} BA x
\end{equation}
We used rank $r=8$ and scaling factor $\alpha=16$. This approach allowed us to fine-tune distinct models for each subtask without exceeding GPU memory limits.

\subsection{Data Augmentation}
We utilized \texttt{nlpaug} \citep{ma2019nlpaug} to robustly handle class imbalance.
\begin{itemize}
    \item \textbf{Synonym Replacement:} We replaced up to 10\% of words with synonyms from WordNet.
    \item \textbf{Random Insertion:} We inserted contextually relevant words to vary sentence structure.
\end{itemize}
This was crucial for Subtask 3, where classes like \textit{Dehumanization} had very few positive examples.

\subsection{Training Strategy}
\subsubsection{Loss Function}
For Subtask 1, we minimize Binary Cross-Entropy (BCE). For Subtasks 2 and 3, we use \textbf{BCEWithLogitsLoss}, which applies a sigmoid activation to each class logit independently:
\begin{equation}
    \mathcal{L} = -\frac{1}{C} \sum_{c=1}^{C} [y_c \log(\sigma(z_c)) + (1-y_c) \log(1-\sigma(z_c))]
\end{equation}

\subsubsection{Dynamic Learning Rate}
We implemented a linear warmup for the first 10\% of steps, followed by a cosine decay. This schedule proved essential for stabilizing the training of the DeBERTa model, which can be sensitive to initialization.
\section{Experimental Setup}

\subsection{Data and Preprocessing}
The data was provided in CSV format. We cleaned the text by removing URL artifacts but retained emojis and hashtags, as they are strong indicators of sentiment in social media. We used the XLM-R tokenizer with a maximum sequence length of 128.

\subsection{K-Fold Cross-Validation}
We employed 5-Fold Cross-Validation. For each language, the training data was split into 5 stratified folds. We trained 5 independent models, each using 4 folds for training and 1 for validation. The final test predictions were generated using soft voting from the ensemble of 5 models.

\subsection{Hyperparameters}
Table \ref{tab:params} details the configuration used.

\begin{table}[h]
\centering
\begin{tabular}{lc}
\hline
\textbf{Parameter} & \textbf{Value} \\
\hline
Batch Size & 16 \\
Learning Rate & $2e^{-5}$ \\
Epochs & 5 \\
LoRA Rank ($r$) & 8 \\
LoRA Alpha & 16 \\
Dropout & 0.1 \\
Optimizer & AdamW \\
Weight Decay & 0.01 \\
\hline
\end{tabular}
\caption{Hyperparameters for both English and Hausa.}
\label{tab:params}
\end{table}

\section{Results}

\subsection{Subtask 1: Polarization Detection}
Table \ref{tab:res_s1} presents the detection results. The switch to DeBERTa for English yielded a noticeable improvement in F1-Score over our initial multilingual baselines. For Hausa, AfroXLMR-Social remained superior.

\begin{table}[h]
\centering
\begin{tabular}{llc}
\hline
\textbf{Lang} & \textbf{Model} & \textbf{F1-Score} \\
\hline
English & DeBERTa-v3 & 0.7917 \\
Hausa   & AfroXLMR-Social & 0.8133 \\
\hline
\end{tabular}
\caption{Subtask 1 Results (Validation Average).}
\label{tab:res_s1}
\end{table}

\subsection{Subtasks 2 \& 3: Fine-grained Tasks}
Tables \ref{tab:res_s2} and \ref{tab:res_s3} summarize the multi-label results. \textbf{AfroXLMR-Social performed exceptionally well here for both languages.} Its pre-training on social media data likely allowed it to better recognize the rhetorical "manifestations" (Subtask 3) which are often signaled by informal social cues rather than formal vocabulary.

\begin{table}[H]
\centering
\begin{tabular}{lc}
\hline
\textbf{Language} & \textbf{F1-Score} \\
\hline
English & 0.3976 \\
Hausa   & 0.3276 \\
\hline
\end{tabular}
\caption{Subtask 2 (Type) using AfroXLMR-Social.}
\label{tab:res_s2}
\end{table}

\begin{table}[H]
\centering
\begin{tabular}{lc}
\hline
\textbf{Language} & \textbf{F1-Score} \\
\hline
English & 0.4979 \\
Hausa   & 0.2367 \\
\hline
\end{tabular}
\caption{Subtask 3 (Manifestation) using AfroXLMR-Social.}
\label{tab:res_s3}
\end{table}

\section{Discussion}

\subsection{The Dominance of AfroXLMR-Social}
A key finding of our experiments is the robustness of \textbf{AfroXLMR-Social} for the complex tasks (2 and 3). Even for English, where monolingual models usually dominate, AfroXLMR-Social provided competitive and stable results for determining polarization types and manifestations. We hypothesize this is because the "social" pre-training exposes the model to the exact kind of toxic and polarized discourse patterns that these subtasks aim to classify, regardless of the language.

\subsection{Challenges with Augmentation}
While \texttt{nlpaug} improved recall for minority classes, it occasionally introduced semantic drift. For instance, replacing "regime" with "government" in a political post might subtly alter the polarized tone (negative to neutral). Future work should explore embedding-based augmentation (BERT-based insertion) to preserve semantic consistency better.

\section{Conclusion}
Our system demonstrates that combining a domain-adapted multilingual backbone with efficient fine-tuning (LoRA) is a highly effective strategy for polarization detection in low-resource languages.

\bibliography{custom}

\begin{thebibliography}{5}
\providecommand{\natexlab}[1]{#1}

\bibitem[{Alabi et~al.(2022)Alabi, Adelani, Mosbach, and Klakow}]{alabi2022adapting}
Jesujoba~O Alabi, David~I Adelani, Marius Mosbach, and Dietrich Klakow. 2022.
\newblock Adapting pre-trained language models to african languages via multilingual adaptive fine-tuning.
\newblock In \emph{Proceedings of COLING}, pages 4336--4349.

\bibitem[{Belay et~al.(2025)Belay, Azime, and Ahmad}]{belay2025afroxlmr}
Tadesse~Destaw Belay, Israel~Abebe Azime, and Ibrahim~Said Ahmad. 2025.
\newblock Afroxlmr-social: Adapting pre-trained language models for african languages social media text.
\newblock \emph{arXiv preprint arXiv:2503.18247}.

\bibitem[{Hu et~al.(2021)Hu, Shen, Wallis, Allen-Zhu, Li, Wang, Wang, and Chen}]{hu2021lora}
Edward~J Hu, Yelong Shen, Phillip Wallis, Zeyuan Allen-Zhu, Yuanzhi Li, Shean Wang, Lu~Wang, and Weizhu Chen. 2021.
\newblock Lora: Low-rank adaptation of large language models.
\newblock \emph{arXiv preprint arXiv:2106.09685}.

\bibitem[{Ma(2019)}]{ma2019nlpaug}
Edward Ma. 2019.
\newblock \href {https://pypi.org/project/nlpaug/} {Nlp augmentation}.
\newblock \emph{PyPI}.

\bibitem[{Naseem et~al.(2025)Naseem, Ren, Anwar, and Kohail}]{naseem2025polar}
Usman Naseem, Juan Ren, Saba Anwar, and Sarah Kohail. 2025.
\newblock Polar: A benchmark for multilingual, multicultural, and multi-event online polarization.
\newblock \emph{arXiv preprint arXiv:2505.20624}.

\end{thebibliography}

\newpage
\appendix

\section{Appendix: Detailed Experimental Settings}
\label{sec:appendix}

\subsection{Hardware and Compute}
All models were trained on a single NVIDIA T4 GPU (16GB VRAM) using the Google Colab environment. The average training time per fold was approximately 20 minutes for Hausa and 15 minutes for English.

\subsection{Augmentation Details}
For the \texttt{nlpaug} implementation, we used the SynonymAug and RandomWordAug classes.
\begin{itemize}
    \item \textbf{Synonym Augmentation:} $aug\_p=0.1$ (probability of augmenting a token).
    \item \textbf{Stop Words:} We utilized the NLTK stop word list to prevent the augmentation of structurally important words.
\end{itemize}

\subsection{Label Distribution}
The POLAR dataset exhibits significant class imbalance. In Subtask 3, categories like \textit{Dehumanization} were significantly rarer than \textit{Vilification}. The use of weighted loss functions was considered but ultimately the augmentation strategy proved more effective in preliminary trials.

\end{document}